\documentclass[10pt,twocolumn,letterpaper]{article}

\usepackage{iccv}
\usepackage{times}
\usepackage{epsfig}
\usepackage{graphicx}
\usepackage{amsmath}
\usepackage{amssymb}
\usepackage{multirow}
\usepackage{booktabs}

\usepackage[breaklinks=true,bookmarks=false]{hyperref}

\iccvfinalcopy 

\def\iccvPaperID{****} 

\ificcvfinal\fi

\begin{document}

\title{VATEX Captioning Challenge 2019:\\Multi-modal Information Fusion and Multi-stage Training Strategy \\for Video Captioning}

\author{Ziqi Zhang\textsuperscript{1,2}\thanks{These authors contributed equally to this work.}, Yaya Shi\textsuperscript{3}\footnotemark[1], Jiutong Wei\textsuperscript{1,2}\footnotemark[1], ChunfengYuan\textsuperscript{1,2}\thanks{Corresponding author.}, Bing Li\textsuperscript{1,2,4}\footnotemark[2], Weiming Hu\textsuperscript{1,2}\\
	\textsuperscript{1}CAS Center for Excellence in Brain Science and Intelligence Technology\\ 
	\textsuperscript{2}National Laboratory of Pattern Recognition, CASIA\\
	\textsuperscript{3}School of Information Science and Technology, University of Science and Technology of China\\
	\textsuperscript{4}PeopleAI, Inc., China\\
	{\tt\small
		\{zhangziqi2017,weijiutong2018\}@ia.ac.cn,shiyaya@mail.ustc.edu.cn,\{cfyuan,bli,wmhu\}@nlpr.ia.ac.cn}
	}
\maketitle
\ificcvfinal\thispagestyle{empty}\fi

\begin{abstract}

Multi-modal information is essential to describe what has happened in a video. In this work, we represent videos by various appearance, motion and audio information guided with video topic. By following multi-stage training strategy, our experiments show steady and significant improvement on the VATEX~\cite{wangvatex} benchmark. This report presents an overview and comparative analysis of our system designed for both Chinese and English tracks on VATEX Captioning Challenge 2019.

\end{abstract}

\section{Introduction}

Video captioning has drawn more attention and shown promising results recently. To translate content-rich video into human language is a extremely complex task, which should not only extract abundant multi-modal information from video but also cross the semantic gap to generate accurate and fluent language. Thanks to the recent developments of useful deep learning frameworks, such as LSTM~\cite{LSTM} networks, as well as of machine translation techniques such as~\cite{seq2seq}, the dominant approach in video captioning is currently based on sequence learning using an encoder-decoder framework. 

In encoding phase, the main task is to well represent given videos general including appearance, motion, audio even speech information. There are many pretrained models can be used to extract above features. In this report, we illustrate our system how to represent videos in detail and use video topic as a global semantic clue to guide better alignment. 

In decoding phase, conventional models follow encoder-decoder framework almost predict the next word conditioned on context information and the previous word. Furthermore, the previous word should be ground truth word at training step but model generated word at inference. As a result, the previous word at training and inference are drawn from different distributions, namely, from the data distribution as opposed to the model distribution. This discrepancy, called \textit{exposure bias}~\cite{exposure} leads to a gap between training and inference. Meanwhile, most models apply cross-entropy loss as their optimization objective, but typically evaluate at inference using discrete and non-differentiable NLP metrics. For above reasons, we apply multi-stage training strategy to train our model to avoid exposure bias problem and directly optimize metrics for the task at hand. Experiments prove that our strategy can obtain steady and significant improvement during training and testing time.

\begin{figure*}[htp]
	\centering
	\includegraphics[width=1\linewidth]{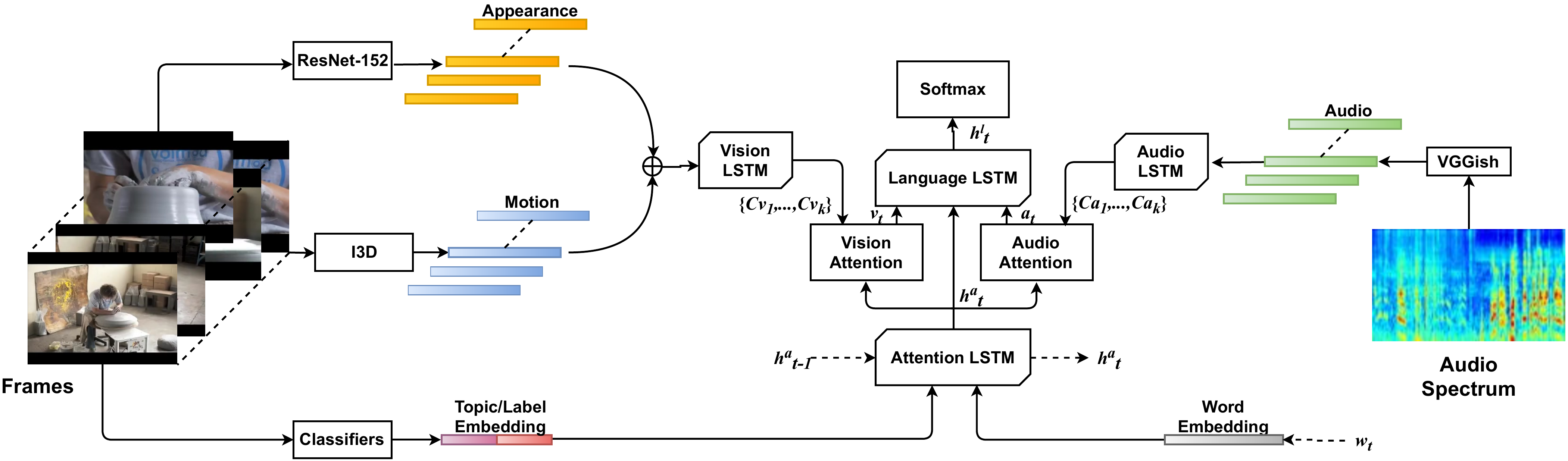}
	\caption[1]{Overview of the proposes Video Captioning framework. Vision LSTM and Audio LSTM are used to selectively merge temporal information of vision features and audio features. And two independent Attention modules can attend to vision and audio with decoding step.}
	\label{fig:1}
\end{figure*}

\section{Multi-modal Video Representations}

We extract the video representations from multiple clues including appearance, motion and audio. We also use video topic to provide global information for specific videos. 

Given one video, we uniformly extract 28 frames as key frames, then select 16 frames around the keyframes as segments. As for those videos whose number of frame less than 28, above selection will be looped. Appearance feature extracted from each frames can reflect global information of these key frames. To extract appearance-based representations from videos, we apply ResNet-152 pretrained on ImageNet dataset. We also attempt some deeper and wider networks \eg InceptionResNet, they barely have improvement for the final result. In order to model the motion information of each segments in video, we use I3D pretrained on Kinetics-600 dataset, which exactly has the same data distribution with VATEX dataset. As for audio feature, though use it alone can not get very good result for video captioning, it can be seen as powerful additional feature, which can provide more discriminative information for some similar content videos. We build audio feature extractor based on VGGish network, which is a variant of the VGG network described in~\cite{vgg}. First, we extract  MEL-spectrogram patches for each of the input audio. The sample rate of the audio is fixed at 16 KHz. The STFT window length is 25 ms and top length is 10 ms. The number of Mel filters is 64. We uniformly sample 28 patches for computing MEL-spectrogram. We then transfer learn from an existing VGGish model which is pretrained on the Audioset dataset~\cite{vggish}. Specifically, we fine-tune this pretrained VGGish model on VATEX training set for 10 epochs. The input size is $ 96\times64 $ for log MEL-spectrogram audio inputs. The last group of convolutional and maxpooling layers are replaced by an embedding layer on the Mel features of size 128. We take this compact embedding layer’s output as our audio feature. In the end, we get $ 28\times2048 $ appearance features, $ 28\times1024 $ motion features and $ 28\times128 $ audio features for each video. Note that each multi-modal feature should be aligned at the same frame to ensure temporal consistency.

Inspired by the Wang's work~\cite{wangtopic}, we found that topic plays an essential role for video captioning. From intuitive understanding, topic can provide global information for specific videos. Topic also can be seen as a cluster, video of the same class always has the similar semantic attributes. We conduct topic-embedding and label-embedding following the same method reported by Wang~\cite{wangtopic}. 

\section{Multi-stage Training Strategy}

In the fist stage, we also apply teacher-forced method to directly optimize the cross-entropy loss. It is necessary to warm-up model during this step. 

In the second step, we utilize word-level oracle method~\cite{wordoracle} to replace conventional scheduled sampling method~\cite{ss}. This method mainly consists of two steps: oracle word selection and sampling with decay. In practice,  by introducing the Gumbel-Max technique we can acquire more robust word-level oracles, which provides a simple and efficient way to sample from a categorical distribution. What's more, the sampling curve is smoother than scheduled sampling method due to its specially designed sampling function. This step can obviously alleviate the problem of overfitting and improve the exploration ability of model. 

It's time to go into the third step when the curve of CIDEr~\cite{cider} metric is no longer growing for 3 epochs. To avoid exposure bias problem, self-critical reinforcement algorithm~\cite{selfcritical} directly optimizes metrics of captioning task. In this work, CIDEr~\cite{cider} and BLEU~\cite{bleu} are equally optimized after the whole sentence generating. This step allow us to more effectively train on non-differentiable metrics, and leads to significant improvements in captioning performance on VATEX.

\begin{table*}[h]
	\begin{center}
		\begin{tabular}{@{}c|cccc|cccc@{}}
			\toprule
			\multirow{2}{*}{Method} & \multicolumn{4}{c|}{English} & \multicolumn{4}{c}{Chinese} \\ 
			& CIDEr $ \uparrow $  & BLEU-4 $ \uparrow $ & ROUGE-L $ \uparrow $ & METEOR $ \uparrow $ 
			& CIDEr $ \uparrow $  & BLEU-4 $ \uparrow $ & ROUGE-L $ \uparrow $ & METEOR $ \uparrow $ \\ 
			\midrule
			Word-level Oracle  & 0.729 & 0.384 & 0.527 & 0.256 & 0.545 & 0.319 & 0.557 & 0.316 \\
			Self-Critical  & \textbf{0.824} & \textbf{0.409} & \textbf{0.542} & \textbf{0.264} 
			& \textbf{0.644} & \textbf{0.326} & \textbf{0.567} & \textbf{0.325} \\
			\bottomrule
		\end{tabular}
	\end{center}
	\caption{Captioning performance of different training strategies on VATEX public test set}
	\label{tab:long}
\end{table*}
\section{Experiments}
\subsection{Dataset}
We utilize the VATEX dataset for video captioning, which contains over 41,250 videos and 825,000 captions in both English and Chinese. Among the captions, there are over 206,000 English-Chinese parallel translation pairs. It covers 600 human activities and a variety of video content. Each video is paired with 10 English and 10 Chinese diverse captions. We follow the official split with 25,991 videos for training, 3,000 videos for validation and 6,000 public test videos for final testing.

\subsection{System}
The overall video captioning framework is illustrated in Figure \ref{fig:1}. In general,  it is composed of two components: 1) Multi-modal video encoder; 2) Top-down~\cite{Anderson2017BottomUpAT} based decoder. In decoding phase, because of the large distribution difference between vision and audio data, we leverage two one-layer LSTM-based architectures to process these two parts of data separately, namely Vision-LSTM and Audio-LSTM. As for vision processing, embedded appearance and motion features are concatenated and then input to Vision-LSTM of 512-D, embedded audio features are directly fed into Audio-LSTM of 128-D. In this way, we can obtain context features with sequential information. During the decoding phase, a top-down based captioning architecture is to be adopted. Attention-LSTM using global video topic and last generated word to guide temporal attention modules to select the most relevant vision and audio regions. Specifically, there are two independent attention modules applying soft-attention to score corresponding regions with topic guide. Meanwhile, Language-LSTM assembles both processed vision and audio context information to generate next word.
\subsection{Evaluations of Video Captioning System}
For this task, four common metrics including BLEU-4, METEOR, CIDEr and ROUGE-L are evaluated. In this subsection, we mainly show steady and significant improvement with different training stage as shown in Table \ref{tab:long}.

\section{Conclusion}
In this report, we explain our designed video captioning system in general. Multi-modal information including appearance, motion and audio are extracted to better represent videos. In order to tackle exposure bias and overfitting problem, we utilize several multi-stage training strategies to train our model. Both Chinese and English tracks are all following the above methods. The experiment proves that our methods can obtain steady and significant captioning performance.

{\small
\bibliographystyle{ieee_fullname}
\bibliography{egbib}
}

\end{document}